\PassOptionsToPackage{hyperfootnotes=false}{hyperref}
\documentclass[pmlr,twocolumn,10pt]{jmlr}

\mlhtrack{findings}

\newif\iffinal
\finaltrue

\iffinal
  \ifmlhneedspmlr
    \jmlrvolume{XXX}
    \jmlryear{2026}
  \fi
  \ifmlhfindings \jmlrproceedings{}{ML4H 2026 - Findings Track}\fi
  \jmlrworkshop{Machine Learning for Health (ML4H) 2026}
\else
  \jmlrproceedings{}{Submitted to ML4H 2026: \mlhtrackname}
  \jmlrworkshop{Machine Learning for Health (ML4H) 2026}
\fi

\hypersetup{
  pdftitle={SurgeCount: A Protocol-Guided Composition Harness for Bimanual Surgical Sponge Counting},
  pdfauthor={Sung Eun Kim, Xiaoman Zhang, Xucheng Wang, Pranav Rajpurkar},
  pdfkeywords={surgical sponge counting, long-horizon manipulation, VLM-orchestrated skill composition, bimanual manipulation}
}
\usepackage{booktabs}
\usepackage{siunitx}
\usepackage{fvextra}
\usepackage{placeins}
\usepackage{tikz}
\usetikzlibrary{positioning,shapes.geometric,arrows.meta,calc,fit,backgrounds}

\newcommand{\equal}[1]{{\hypersetup{linkcolor=black}\thanks{#1}}}

\title[SurgeCount]{SurgeCount: A Protocol-Guided Composition Harness for Bimanual Surgical Sponge Counting}
\author[Kim, Zhang, Wang, and Rajpurkar]{%
\Name{Sung Eun Kim}\equal{These authors contributed equally.},
\Name{Xiaoman Zhang}\footnotemark[1],
\Name{Xucheng Wang}\footnotemark[1], and
\Name{Pranav Rajpurkar}\\
\addr Department of Biomedical Informatics, Harvard Medical School, Boston, MA, USA\\
\addr Correspondence: Sung Eun Kim, \href{mailto:sungeun_kim2@hms.harvard.edu}{\nolinkurl{sungeun_kim2@hms.harvard.edu}}
}

\begin{document}
\maketitle
\begin{figure*}[t]
\centering
\includegraphics[width=0.85\linewidth]{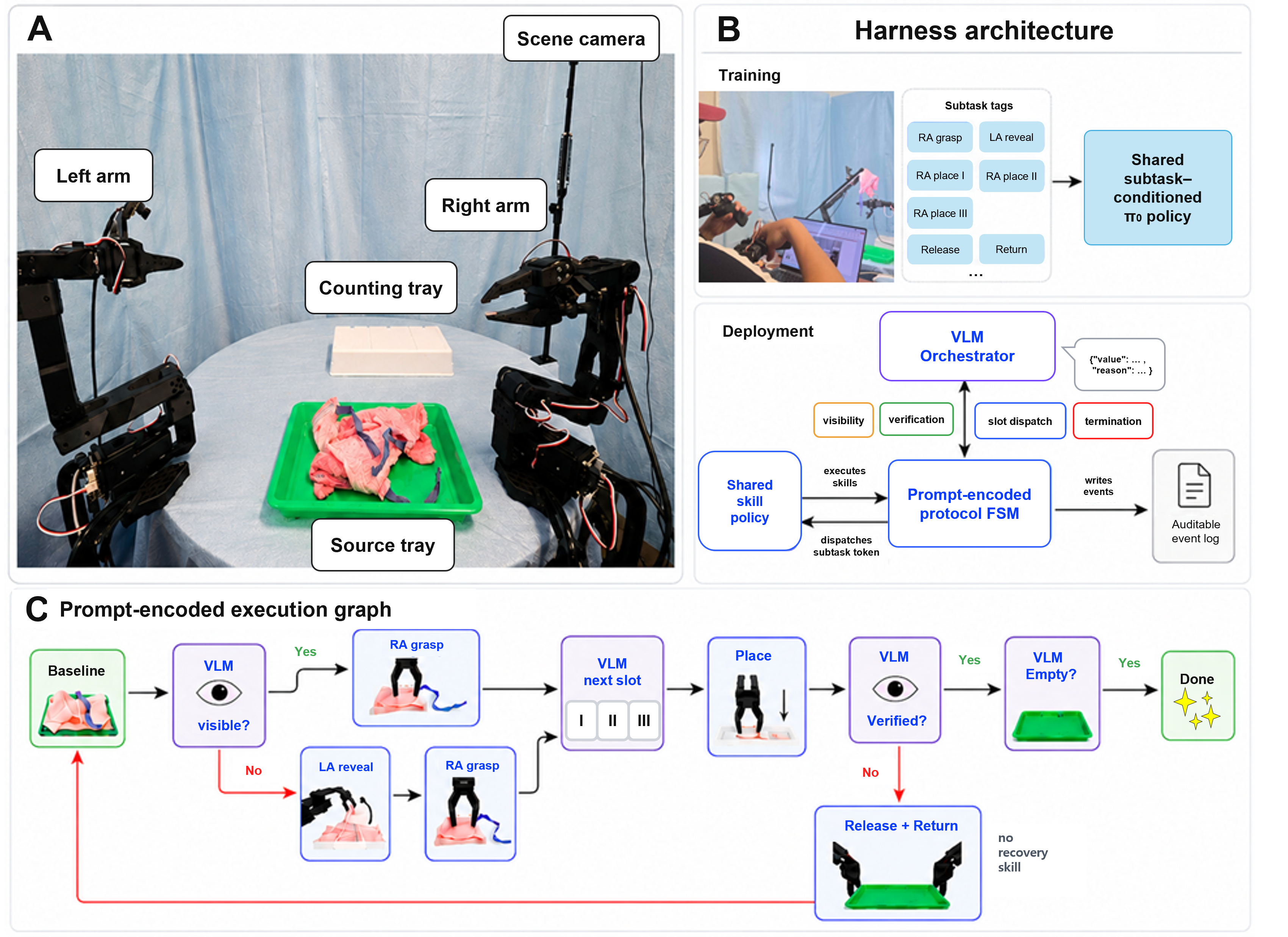}
\caption{\textbf{SurgeCount system overview.}
\textbf{(A)} Two SO-101 arms with wrist cameras and a fixed scene camera move red-stained sponge from the source tray (sponge tray) to a three-slot counting tray, using the blue radiopaque thread as the preferred isolation landmark.
\textbf{(B)} A subtask-conditioned $\pi_0$ policy with a learned stop head executes one subtask at a time; at each stop, a frozen VLM selects the next deployment command under a prompt-encoded protocol finite-state machine (FSM), logging every decision.
\textbf{(C)} The orchestrator routes between a visible-thread and an occluded-thread (left-arm reveal) strategy; both converge on placement and verification, with gate failures triggering composed release-and-return recovery.
}
\label{fig:overview}
\end{figure*}

\begin{abstract}
We present SurgeCount, a system for auditable robotic surgical sponge counting that composes learned skills under a written procedural protocol. Retained sponges cause preventable harm, and surgical teams count every sponge before closure to reduce that risk. The robot repeatedly isolates one deformable sponge from a tangled pile and places it in the next numbered counting-tray slot, requiring handling of occluded grasp threads while preserving step order. We train one subtask-conditioned policy on eleven short skills and learn when each skill ends. At each learned boundary, a stateful vision-language model selects the next skill from an operator-authored specification of step order, visual verification, and recovery. This creates temporal commitment to each subtask and enables failures to be recovered by recomposing existing skills rather than using a dedicated recovery policy. We evaluate five methods across 100 real-robot trials using 20 matched layouts. SurgeCount fills 47 of 60 ordered slots and completes 12 of 20 procedures, compared with 27 of 60 and 3 of 20 for the strongest of four end-to-end policies trained on the same data. Replacing the detailed protocol prompt with a minimal prompt reduces performance to 29 of 60 slots and 5 of 20 completed procedures. These results show that long-horizon performance depends not only on visuomotor capacity, but also on how learned skills are committed, verified, and composed. This tabletop proof of concept motivates further evaluation of protocol-guided robotic sponge counting in clinically representative settings.
\end{abstract}


\ifmlhneedsstatements
\paragraph*{Data and Code Availability}
Code for data collection, preparation, policy training, and deployment with VLM orchestration, together with representative execution and recovery videos, is available at \url{https://github.com/rajpurkarlab/SurgeCount}. The supplementary appendix includes orchestrator prompts and additional experimental details. 

\paragraph*{Institutional Review Board (IRB)}
This robot-only study involved no human participants, patient data, biological specimens, or animals. Institutional Review Board review was therefore not applicable.
\fi

\section{Introduction}

Retained surgical sponges can cause serious adverse events, and prevention depends on accounting for every sponge before wound closure~\citep{gawande2003retained}. Automating such a workflow poses a broader challenge for learned healthcare robotics: procedural competence requires not only accurate visuomotor actions, but also maintaining step order, committing to a subtask until completion, verifying its outcome, and recovering when an imperfect skill fails.

End-to-end vision-language-action (VLA) policies typically absorb these functions into a continuous sequence of predictions. For long-horizon procedures, however, hierarchical prediction alone may be insufficient if the policy repeatedly reconsiders its active subtask before completion. In our evaluation, the hierarchy-aware $\pi_{0.5}$ model re-issues its subtask prediction before each action chunk and frequently changes that prediction between consecutive chunks, despite operating on the same underlying visuomotor backbone as $\pi_0$. This motivates a different formulation: rather than asking a single policy to infer procedural structure implicitly, learned policies can execute bounded skills while explicit procedural knowledge determines when transitions are allowed, how outcomes are verified, and how failures are recovered.

We study this question through robotic surgical sponge counting. Three red-ink-stained surgical sponges overlap in a source tray, and each carries a blue radiopaque thread that serves as the preferred grasp landmark for isolating a single sponge. The robot must transfer one sponge at a time into counting-tray slots I, II, and III in order. When the thread is visible, the right arm can grasp it directly; when it is occluded by overlapping sponges, the left arm must first manipulate the pile to expose the thread before the right arm can grasp and transfer the isolated sponge. The task therefore couples deformable-object manipulation, visual ambiguity, bimanual coordination, ordered placement, and repeated verification.

We introduce \textbf{SurgeCount}, a protocol-guided composition harness over an off-the-shelf $\pi_0$ policy that externalizes procedural hierarchy through three coupled mechanisms (Figure~\ref{fig:overview}). First, a single subtask-conditioned $\pi_0$ checkpoint executes eleven learned skills, and a learned stop head establishes \textbf{temporal commitment}: once a skill is selected, it runs until a learned completion boundary rather than being re-decided at every action chunk. Second, only at these boundaries does a stateful vision-language model select the next skill under an operator-authored protocol specifying legal orderings, two nominal strategies (visible-thread grasp and occluded-thread reveal-then-grasp), and per-skill visual verification gates. Third, when a gate fails, recovery is achieved by recomposing release-and-return subtasks already present in the learned skill library, without a dedicated recovery policy~\citep{dai2024racer}. Representative rollouts, including the bimanual reveal and composed recovery, are shown in the supplementary videos at \url{https://github.com/rajpurkarlab/SurgeCount}.

This design deliberately preserves procedural knowledge as an explicit supervisory layer rather than requiring it to be rediscovered entirely from robot demonstrations. Beyond the shared teleoperation data, SurgeCount uses operator-entered subtask tags, stop labels derived from those segmentations, and an operator-authored protocol specification; it uses no transition-level reasoning annotations, no learned recovery policy, and no engineered affordance API. Every orchestrator transition, verification decision, and recovery step is logged, making failures localizable at explicit subtask boundaries and the resulting execution trace directly auditable.

\paragraph{Contributions.}
\begin{itemize}
    \item \textbf{Protocol-guided skill composition}: a single subtask-conditioned VLA, learned completion boundaries, and a stateful VLM constrained by operator-authored ordering and verification rules.
    \item \textbf{Temporal commitment and compositional recovery}: learned boundaries prevent per-chunk subtask re-decoding and provide explicit points for verification and recovery using already-trained skills.
    \item \textbf{Real-robot evaluation}: 100 matched-layout trials against four end-to-end policies, with mechanistic diagnostics and a protocol-specificity ablation.
\end{itemize}

\section{Related Work}
\label{sec:relatedwork}

\paragraph{Long-Horizon VLA Composition.}
Behavior cloning supports short-horizon manipulation~\citep{zhao2023learning,chi2024diffusion,black2024pi0,intelligence2025pi05,shukor2025smolvla,kim2024openvla,brohan2023rt2}, but long-horizon procedures also require subtask boundaries, verification, and recovery. Prior approaches use large-scale pretraining~\citep{intelligence2025pi05}, per-transition reasoning annotations~\citep{lin2025onetwovla,zawalski2024ecot}, dedicated recovery policies~\citep{dai2024racer,ablett2020fire,gokmen2023bcva}, or structured skill composition~\citep{sutton1999options,garg2022lisa,ahn2022saycan,liang2022codeaspolicies}. SurgeCount instead combines subtask-tagged demonstrations with a written protocol and recovers by recomposing existing skills, without transition-level reasoning annotations or a separate recovery policy.

\paragraph{Bimanual and Surgical Manipulation.}
Our SO-101 setup follows the ALOHA bimanual lineage~\citep{fu2024mobile,aldaco2024aloha2}. Related work uses additional viewpoints, stabilization, or data augmentation~\citep{he2026bimanual,zeng2025activeumi,xiong2025via,shen2025dcoda}; SurgeCount uses the second arm to reveal an occluded grasp landmark. Surgical robotics spans teleoperation~\citep{kazanzides2014dvrk}, laparoscopic subtasks~\citep{scheikl2023lapgym,saeidi2022star}, open-surgery suturing~\citep{wang2026opensurgery}, language-conditioned assistance~\citep{li2024robonurse}, and object detection~\citep{lai2026detection}. We study ordered sponge counting with verification and recovery.

\paragraph{VLM-Orchestrated Learned Skills.}
Prior systems use frozen language or vision-language models to orchestrate skills through primitives or engineered interfaces~\citep{ahn2022saycan,huang2022inner,liang2022codeaspolicies}. SurgeCount combines a single learned subtask-conditioned $\pi_0$ policy, learned stop boundaries, and a prompt-encoded protocol for selecting trained skills or the non-learned \texttt{ra\_home} and \texttt{wait} commands. Unlike end-of-episode success detection~\citep{du2023successvqa,kang2025robomonkey}, verification occurs at every learned boundary to guide subsequent execution.

\section{Method}
\label{sec:method}

\subsection{Task and System Overview}
\label{sec:method-problem}

SurgeCount externalizes hierarchy through commitment, stateful selection, and outcome enforcement. Its learned skill library contains eleven single-arm-active subtasks executed by one shared policy checkpoint (Table~\ref{tab:dataset}). The deployment interface extends these learned tags with two non-learned commands: \texttt{ra\_home}, a hardcoded right-arm return, and \texttt{wait}, which holds the current pose without dispatching a manipulation skill.

SurgeCount has two components. A subtask-conditioned $\pi_0$ policy executes the learned skills from three RGB views and both arms' proprioception, producing 12-dimensional bimanual actions. A stateful VLM orchestrator selects the next learned tag or non-learned command according to legal orderings, verification gates, and recovery rules encoded in its prompt. The policy runs a selected learned subtask until its stop signal fires; the controller then freezes the arms and queries the orchestrator with the stopped frames, compact state, and history. A returned learned tag resumes policy execution, \texttt{ra\_home} invokes the hardcoded return, and \texttt{wait} preserves the pose pending another observation. The trial terminates when the source tray is empty or a timeout condition is reached (\S\ref{sec:method-fsm}); \texttt{wait} only defers robot action.

\subsection{Subtask-Conditioned \texorpdfstring{$\pi_0$}{pi0} Policy with Stop-Signal Boundaries}
\label{sec:method-policy}

\paragraph{Policy and Subtask Conditioning.}
We adopt the $\pi_0$ VLA backbone ~\citep{black2024pi0}, a PaliGemma vision/language tower with a Gemma action expert trained as a flow-matching policy. 
Subtask conditioning routes through the existing language pathway: each tag carries a short language description that the $\pi_0$ pre-processor formats as a prompt (tokenizer and prompt details in Appendix~\ref{app:repro}). At deployment, the controller swaps the description string on each transition, so changing subtasks is a token-stream edit.

\paragraph{Learned Stop Signal.}
We keep the $\pi_0$ backbone unchanged and add a single scalar stop head to predict subtask completion. Stop labels are derived automatically from the existing teleoperation segmentation: the final 10 frames of each subtask are labeled $\texttt{stop}=1$, and all earlier frames $\texttt{stop}=0$. At deployment, the controller declares the subtask complete when the policy emits a sustained stop signal for 10 consecutive control steps, then freezes the arms at the stopped pose while the orchestrator selects the next subtask. Boundary detection is therefore learned from the same teleoperation data, without periodic VLM polling or hand-tuned action-overflow tolerances.

\paragraph{Single-Arm-Active Training Hooks.}
Each learned subtask commands only one arm, so the idle arm's targets are static and learnable from proprioception alone. Two data-loader hooks exploit this: per-subtask camera masking zeros non-active camera inputs before the vision encoder, and idle-frame filtering drops frames recorded with the \texttt{wait} tag from gradient updates. The recorded \texttt{wait} tag denotes an idle/no-op interval, not a learned manipulation skill. Neither hook modifies the $\pi_0$ flow-matching loss over all 12 action dimensions. A single checkpoint is trained jointly across all eleven tags and reused for every dispatch (Figure~\ref{fig:overview}B; implementation in Appendix~\ref{app:repro}).

\paragraph{Corrective Recovery Data via Offline DAgger.}
Expert teleoperation alone leaves each subtask brittle at the out-of-distribution states it reaches under its own errors. Following the DAgger principle of training on the learner's own state distribution~\citep{ross2011dagger,xu2025crdagger}, we run an offline aggregation loop: we roll out the current checkpoint and, wherever it begins to fail, an operator teleoperates a short corrective trajectory back to a recoverable configuration; the policy's failing actions are discarded and only the corrective segments are kept, aggregated with the expert demonstrations, and used to fine-tune the same $\pi_0$ checkpoint, with no separate recovery network. These corrective segments are tagged like any other subtask and are part of the shared demonstration set used by every method (\S\ref{sec:protocol}); they harden each low-level subtask against its own error states, which is distinct from the orchestrator-level recovery of \S\ref{sec:method-fsm}.

\subsection{Prompt-Encoded Protocol FSM and VLM Skill Orchestration}
\label{sec:method-verifier}

One operator-authored prompt specifies the workspace, learned subtask tags, two nominal strategies, preferred protocol-consistent action sequences, per-skill visual verification, recovery instructions, and terminal behavior. The VLM acts only as a high-level dispatcher. Unlike free-form planning~\citep{huang2022inner}, it neither generates joint-level actions nor introduces skills outside the trained library. Its structured response is restricted to the eleven learned tags, \texttt{ra\_home}, and \texttt{wait}. Together, the prompt, response schema, verification gates, and recovery logic constrain VLM decisions to the trained skill interface and mitigate semantically incorrect selections.

\paragraph{Orchestrator Input and Output.}
At each boundary, the controller queries an external VLM~\citep{du2023successvqa,kang2025robomonkey} with a fixed contract: three RGB frames at the stopped pose, a compact JSON state (last subtask, inferred slot-fill status), and the chat history of prior decisions in the same episode. The VLM returns one of the eleven learned tags, \texttt{ra\_home}, or \texttt{wait}, together with a one-sentence rationale and scalar blue-thread visibility confidence; all outputs are logged for analysis. No ground-truth evaluation labels enter the contract.

\paragraph{Verification Gates.}
The prompt defines a visual check per skill. The direct-grasp gate confirms that the blue thread is held. In the reveal strategy, the left arm first isolates one sponge, and the subsequent gate accepts a right-arm hold on that isolated sponge by either thread or fabric. The placement gate confirms slot arrival before release; the baseline gate confirms tray emptiness and, when satisfied, selects \texttt{wait}. These checks are the sole source of slot-fill state; the orchestrator never reads evaluation labels.

\paragraph{Conditional Prompt Injections.}
Two pieces of episode state are appended when relevant: a wrong-grab counter that triggers a switch to the occluded-thread strategy after consecutive failed grasps, and timeout context that injects recovery guidance on a stall (\S\ref{sec:method-fsm}). Full templates are in Appendix~\ref{app:verifier-prompt}.

\paragraph{Implementation.}
We use Gemini 3.1 Pro for all VLM queries, selected for its
structured-output JSON support and per-subtask diagnostic
accuracy among candidates (Appendix~\ref{app:verifier-acc}).
Each robot episode uses one continuous VLM chat; all tags,
rationales, confidences, and transitions are logged for
post-hoc audit.
Across 481 VLM calls during the main robot evaluation,
mean response latency was 8.85\,s
(sample SD, 5.31\,s; median, 7.31\,s).
These measurements reflect deployment queries with the
procedural prompt and accumulated episode context.

\subsection{Protocol Strategies, Recovery, and Termination}
\label{sec:method-fsm}

The prompt names two happy-path strategies (Figure~\ref{fig:overview}C), each a sequence over the learned skill library (Table~\ref{tab:dataset}) that the orchestrator follows unless a gate fails, sharing the place-release-return tail $2x \to 3 \to 4$ (where $2x$ is the next empty slot, \texttt{2a}/\texttt{2b}/\texttt{2c}). The \emph{visible-thread} strategy uses the right arm alone, $1 \to 2x \to 3 \to 4$. The \emph{occluded-thread} strategy first has the left arm lift a neighboring gauze's white edge to expose the thread for the right arm, giving $5 \to 6 \to 7 \to 9 \to 2x \to 3 \to 4$ (Appendix~\ref{app:graspseq} details this reveal-then-grasp sequence). The end-to-end policy moves both arms but does not chain reveal-then-grasp reliably; the named sequences and per-skill gates enforce the ordering.

\paragraph{Recovery as Orchestrator Composition.}
Recovery is a property of this design rather than a separately engineered mechanism: because the orchestrator already composes the procedure from existing subtasks under verification gates, a gate failure is handled by composing those same subtasks, with no dedicated recovery policy~\citep{dai2024racer}. Procedure-level recovery is recomposition of existing subtasks, while the subtasks' own robustness to error states comes from the corrective data of \S\ref{sec:method-policy}. For example, if the right-arm gripper does not hold the blue thread after tag~1 or tag~6, the orchestrator returns the arm to baseline using release-and-return subtasks already in the training set, then resumes from a fresh visibility assessment. Every recovery transition is logged like any other orchestrator decision, so the recovery path is auditable. A hardcoded right-arm home primitive serves as a safety fallback, tracked separately from learned-subtask recovery.

\paragraph{Termination and Timeout.}
Correct slot filling is an outcome metric, not the stopping rule, because a run can empty the source tray while placing gauze incorrectly. SurgeCount's 1-minute per-subtask stall timeout (available because learned boundaries exist) injects timeout context that either triggers recovery or terminates the trial; baseline timeouts are specified in \S\ref{sec:protocol}.

\section{Experiments}
\label{sec:experiments}

\subsection{Hardware and Task Setup}
\label{sec:hardware}

\paragraph{Platform and Observations.}
We evaluate on two small-form-factor SO-101 LeRobot arms in a leader--follower configuration~\citep{fu2024mobile,aldaco2024aloha2}, denoted RA (right) and LA (left) and observed by three RGB cameras (right-wrist, left-wrist, and a fixed overhead scene camera). The low-cost SO-101 exhibits non-trivial joint-angle noise and gripper backlash, which the learned policy must absorb without explicit compensation; capture and control rates are in Appendix~\ref{app:repro}.

\paragraph{Workspace and Surgical Sponges.}
A green source tray and a white three-slot counting tray (I, II, III) sit between the arms (Figure~\ref{fig:overview}A). Each item is a 12 $\times$ 12 inch (30 $\times$ 30 cm) surgical sponge with a protruding blue radiopaque thread, and red ink used as a reproducible visual proxy. The thread is the preferred landmark for isolating one sponge directly from the tangled pile because a direct body grasp risks lifting several and corrupting the count. After the left arm has isolated one sponge during the reveal strategy, however, the right arm may transfer that sponge by either thread or fabric. Each trial begins with three overlapping sponges, requiring the full I$\to$II$\to$III workflow.

\subsection{Dataset and Demonstration Protocol}
\label{sec:dataset}

Demonstrations are collected by leader--follower teleoperation following the two protocol strategies of \S\ref{sec:method-fsm}. Each episode is tagged at recording time with the active subtask, from which stop-signal labels are derived automatically (\S\ref{sec:method-policy}).

The data loader filters the \texttt{wait} token during policy training, leaving eleven single-arm-active manipulation subtasks (per-subtask counts in Appendix~\ref{app:dataset}). In total, the training set contains 2{,}066 episodes and 158{,}771 frames across the eleven trained subtasks; an additional 276 episodes and 23{,}077 frames carrying the \texttt{wait} token are filtered out and not used for policy training.

\subsection{Baselines and Evaluation Protocol}
\label{sec:protocol}

\paragraph{Baselines.}
We compare four end-to-end baselines against SurgeCount (Table~\ref{tab:baselines}). We include $\pi_{0.5}$ as a hierarchy-aware end-to-end baseline with reactive subtask decoding. SurgeCount is orchestrated at deployment by Gemini 3.1 Pro, chosen because it is by far the most accurate at naming the next subtask from a single multi-view still (Appendix~\ref{app:verifier-acc}).

\begin{table*}[t]
\centering
\footnotesize
\caption{Methods compared in the main evaluation.}
\label{tab:baselines}
\begin{tabular}{ll}
\toprule
Method & Notes \\
\midrule
ACT~\citep{zhao2023learning}             & Flat policy, no subtask structure \\
SmolVLA~\citep{shukor2025smolvla}        & Flat VLA, no subtask structure \\
$\pi_0$~\citep{black2024pi0}             & Flat VLA, no subtask structure \\
$\pi_{0.5}$~\citep{intelligence2025pi05} & Built-in subtask tokens, re-decoded every chunk (reactive) \\
\textbf{SurgeCount (ours)}               & Subtask-conditioned $\pi_0$ + learned stop + VLM orchestrator (committed) \\
\bottomrule
\end{tabular}
\end{table*}

\paragraph{Evaluation Protocol.}
Each method is evaluated on 20 trials of the three-gauze counting procedure, split evenly between visible-thread and occluded-thread initial configurations. Starting configurations are paired across methods using photographed reference layouts: for each of the 20 trials, we record the initial source-tray configuration and manually reproduce the same visible-thread or occluded-thread layout as closely as possible for every method. The full set of reference configurations is shown in Appendix~\ref{app:configs}. End-to-end baselines are continuous rollouts without subtask boundaries or recovery dispatch; they are stopped after a 5-minute trial-level timeout if they remain stuck in a loop or continue without meaningful progress. SurgeCount uses learned subtask boundaries and therefore applies a 1-minute per-subtask stall timeout, which can trigger orchestrator-dispatched recovery before the trial ends (\S\ref{sec:method-fsm}).

\paragraph{Trial Cost.}
Reproducing the 20 initial configurations across all five methods totals 100 paired rollouts (120 with the ablation, $\sim$10 hours of robot time); the matched-layout protocol gives paired comparisons, raising per-trial statistical power over unpaired evaluation.

\paragraph{Endpoint Metrics.}
We report two endpoint metrics. First, \emph{slots filled} is the total number of sponges correctly placed into the next expected ordered slot, out of $20 \text{ trials} \times 3 \text{ slots} = 60$ ordered slots. A slot is counted as correct only when exactly one sponge is isolated from the source pile and placed into the next expected slot in I$\to$II$\to$III order. Scoring therefore depends on single-sponge isolation and ordered placement, not on whether the final grasp contacts the thread or fabric; the blue thread is the preferred direct-grasp landmark because it reduces the risk of lifting multiple sponges. Out-of-order placements, multi-sponge transfers, or placements into the wrong slot do not count. Second, \emph{full-task success} is the number of trials in which all three slots are correctly filled in order, out of 20. These metrics are computed independently of the stopping rule: a trial can terminate because the source tray is empty while still receiving partial or zero slot credit if the placements were incorrect.

\paragraph{Ablations.}
We report a prompt-specificity ablation comparing the full protocol prompt against a minimal prompt with reduced task-specific verification and recovery guidance. It uses the same hardware, dataset, and endpoint metrics as the main evaluation, with each condition run for 20 matched trials. We assess significance with two-sided Wilcoxon signed-rank tests on per-trial slot counts (20 paired observations, each in $\{0,1,2,3\}$) and McNemar's exact tests on paired procedure success; because slots within a trial are not independent, we do not test the 60-slot marginals with Fisher's exact.

\section{Results}
\label{sec:results}

\subsection{Temporal Commitment Improves Long-Horizon Execution beyond Hierarchical Prediction}

Among the five evaluated methods, SurgeCount achieved the highest ordered-placement and full-procedure completion rates, filling 47 of 60 ordered slots and completing 12 of 20 procedures (Table~\ref{tab:main}). The strongest end-to-end baseline, $\pi_0$, achieved 27 of 60 slots and 3 of 20 complete procedures.

\begin{table}[t]
\centering
\caption{Main comparison across 20 matched-layout trials per method (10 visible-thread, 10 occluded-thread). \emph{Partial-task} reports correctly filled ordered slots out of 60; \emph{full-task} reports trials with all three slots correctly filled in order out of 20.}
\label{tab:main}
\footnotesize
\begin{tabular}{lcc}
\toprule
Method & Partial-task & Full-task \\
\midrule
ACT & 7 / 60 & 0 / 20 \\
SmolVLA & 12 / 60 & 0 / 20 \\
$\pi_0$ & 27 / 60 & 3 / 20 \\
$\pi_{0.5}$ (E2E) & 15 / 60 & 2 / 20 \\
\textbf{SurgeCount (ours)} & \textbf{47 / 60} & \textbf{12 / 20} \\
\bottomrule
\end{tabular}
\end{table}

The comparison with $\pi_{0.5}$ is especially informative. Although $\pi_{0.5}$ shares the $\pi_0$ lineage while adding large-scale language pretraining and hierarchical subtask prediction, it achieves only 15 of 60 slots and 2 of 20 complete procedures. The distinction is therefore not simply whether hierarchy is represented, but how hierarchy is executed. $\pi_{0.5}$ re-decodes its active subtask before each action chunk, whereas SurgeCount commits to the selected subtask until a learned completion boundary and transitions only after outcome verification.

Across 20 rollouts, $\pi_{0.5}$ changes its predicted active subtask between consecutive action chunks 18.2 times per trial on average, compared with 0 within a committed SurgeCount subtask by construction. These switches occur most often at the reveal-to-grasp transition, where $\pi_{0.5}$ oscillates between the left-arm reveal and its return rather than completing the right-arm exposed-thread grasp; it also emits degenerate subtask strings (e.g. ``gragra...'') on 2.5\% of decodes. These observations indicate that reactive hierarchical prediction is not equivalent to temporal abstraction: long-horizon execution benefits from explicit commitment to subtask completion.

Across paired layouts, SurgeCount's slot fill exceeds the strongest end-to-end baseline, $\pi_0$ (Wilcoxon signed-rank, $p<0.01$), and the procedure-completion gap is significant (McNemar's exact, $p<0.05$).
% Exact paired tests, Ours vs strongest baseline pi0 (Ours slots [3,3,2,3,1,2,2,3,0,3,3,1,3,3,3,1,3,2,3,3]=47/60, 12/20; pi0 [1,1,2,3,3,1,2,0,1,1,0,2,1,3,2,1,0,0,1,2]=27/60, 3/20): Wilcoxon signed-rank p approx 0.009 (two-sided), McNemar exact p approx 0.012 (two-sided; b=10 Ours-only successes, c=1 pi0-only). Ablation full-vs-minimal: Wilcoxon two-sided p approx 0.008, McNemar two-sided p approx 0.065 (b=9, c=2). Recompute if pi0/pi0.5/ACT/SmolVLA are re-tested.

\subsection{Verification and Retry Convert Imperfect Skills into Procedure-Level Success}
\label{sec:results-harness}

SurgeCount does not succeed because every learned manipulation skill is reliable. The direct visible-thread grasp succeeds on 53.6\% of attempts and the exposed-thread grasp after reveal on only 32.7\%, whereas placement and release-and-return subtasks succeed on every observed attempt once the sponge has been isolated (Table~\ref{tab:ours-diagnostics}).

\begin{table*}[t]
\centering
\caption{Per-attempt SurgeCount diagnostics across 20 trials. \emph{Trials invoked} counts how many trials called each subtask group at least once. \emph{Per-attempt success} is the fraction of attempts that complete the subtask successfully. Tag IDs reference Table~\ref{tab:dataset}.}
\label{tab:ours-diagnostics}
\footnotesize
\begin{tabular}{llcr@{\,/\,}lr}
\toprule
Subtask group & Tag(s) & Trials invoked & \multicolumn{2}{c}{Accepted\,/\,attempts} & Rate \\
\midrule
RA grasp visible thread     & \texttt{1}           & 20 / 20 & 37  & 69  & 53.6\% \\
LA reveal (lift white edge) & \texttt{5}           & 17 / 20 & 69  & 88  & 78.4\% \\
RA grasp exposed thread     & \texttt{6}           & 16 / 20 & 17  & 52  & 32.7\% \\
RA place in slot            & \texttt{2a, 2b, 2c}  & 19 / 20 & 50  & 50  & 100\% \\
RA release + return         & \texttt{3, 4}        & 20 / 20 & 112 & 112 & 100\% \\
LA release + return         & \texttt{7, 8, 9}     & 19 / 20 & 110 & 110 & 100\% \\
\bottomrule
\end{tabular}
\end{table*}

Despite this low per-attempt grasp reliability, SurgeCount completes 12 of 20 full procedures. The mechanism is failure containment: unsuccessful grasps are surfaced at explicit subtask boundaries, verified as unsuccessful, followed by release-and-return, and retried from a recoverable state rather than propagated into the next placement (Appendix~\ref{app:failures} catalogs the observed failure modes). SurgeCount averages 4.3 retries per trial, with 77.9\% occurring on the two thread-grasp subtasks; the occluded-thread reveal yields at least one successful exposed-thread grasp in 12 of the 17 trials that invoke it.

The orchestrator misjudged a completed subtask 8 times across 481 executions (1.7\%), and each event was logged. Thus, the composition layer does not eliminate the dominant low-level failure mode; it changes what happens after failure. \textit{The harness does not make grasping easy; it makes failed grasps recoverable and auditable.}

\subsection{Explicit Procedural Guidance Contributes beyond the Learned Skill Library}
\label{sec:ablation}

To assess the contribution of detailed protocol guidance beyond the learned skill library, we compare the full protocol prompt against a minimal prompt that retains the same skill library and task goal, including basic preconditions, ordering instructions, and retry guidance, but omits the two named strategies and detailed verification and recovery rules (verbatim in Appendix~\ref{app:verifier-prompt}). All other components are held fixed.

Across 20 matched trials per condition, the full protocol reaches 47 of 60 ordered slots and 12 of 20 complete procedures, whereas the minimal prompt reaches 29 of 60 and 5 of 20, respectively (Wilcoxon, $p<0.01$ on partial-task; McNemar, $p\approx0.07$ on full-task). Removing explicit procedural guidance also degrades the orchestrator itself: its rate of misreading a completed subtask rises from 1.7\% under the full protocol (\S\ref{sec:results-harness}) to 10.6\% under the minimal prompt, a sixfold increase.

These results indicate that task decomposition alone is insufficient. A task goal and learned skill vocabulary do not reliably recover the procedural constraints required for execution. The operator-authored protocol contributes explicit knowledge about when a skill is valid, what successful completion looks like, and how failure should alter the next decision.

\section{Conclusion}
\label{sec:conclusion}

SurgeCount shows that long-horizon bimanual procedures can improve without redesigning the underlying VLA backbone. Over an off-the-shelf $\pi_0$ policy, learned subtask boundaries and protocol-constrained VLM orchestration improve ordered placement and procedure completion relative to four tested end-to-end policies. The results suggest that long-horizon performance depends not only on visuomotor capacity, but also on how learned skills are temporally committed, verified, and composed.

The system's strongest gains arise not from eliminating low-level failures, but from changing what happens after them. Failed grasps are surfaced at explicit boundaries, returned to a recoverable state, and retried rather than propagated into subsequent steps. The protocol ablation further shows that a task goal and learned skill vocabulary alone do not recover the procedural constraints required for reliable execution. Together, these findings support explicit procedural structure as a useful complement to learned visuomotor policies for auditable long-horizon workflows.

The intended use case is terminal sponge reconciliation during surgical closure and sign-out, rather than manipulation within a rapidly changing operative field. Cloud-query latency therefore affects overall workflow duration rather than delaying time-critical tissue manipulation.
\paragraph{Interpretation of the Comparisons.}
The comparison does not isolate the contribution of temporal commitment: SurgeCount also adds a VLM orchestrator, an operator-authored protocol, and verification and recovery mechanisms. The prompt ablation demonstrates a benefit of the detailed prompt for ordered-slot performance, but does not separate the effects of ordering rules, verification criteria, and recovery guidance. The procedure-completion difference between prompt conditions did not reach statistical significance at the 0.05 level ($p\approx0.07$).
\paragraph{Limitations.}\label{sec:limitations}
This study is a tabletop proof of concept involving one fixed procedural workflow. The dominant failure mode, unreliable thread grasping, is retried rather than eliminated (Appendix~\ref{app:failures}). We do not establish robustness to changes in lighting, camera placement, tray geometry, sponge appearance, or workspace configuration, nor transfer to other procedural tasks or tightly coupled continuous-contact manipulation. Future work should test the framework across additional healthcare workflows and environments.

\bibliography{references}
\clearpage

% Appendix - included from main.tex AFTER \bibliography
\appendix

{\centering \Large\bfseries Appendix\par}
\vspace{0.75em}

% Full-width demonstration-count table is declared here, at the top of the
% first appendix page, so that it is placed on this page rather than being
% deferred to the next page and stranding the Appendix B heading.
\begin{table*}[!t]
\centering
\caption{Per-subtask demonstration counts (referenced from \S\ref{sec:dataset} and Appendix~\ref{app:dataset}). Tag IDs follow the data-collection protocol; rows are listed in protocol-flow order. The \texttt{wait} token is present in the recordings but filtered by the data loader and not used as a learned manipulation skill.}
\label{tab:dataset}
\begin{tabular}{llllr}
\toprule
Tag & Arm & Subtask & Episodes & Frames \\
\midrule
\texttt{1}  & RA & Grasp visible blue thread from source tray & 239 & 36,145 \\
\texttt{5}  & LA & Grasp white edge for reveal                & 219 & 28,462 \\
\texttt{6}  & RA & Grasp exposed blue thread                  & 242 & 36,833 \\
\texttt{2a} & RA & Place gauze in slot I                      & 101 & 4,202 \\
\texttt{2b} & RA & Place gauze in slot II                     & 100 & 3,828 \\
\texttt{2c} & RA & Place gauze in slot III                    & 100 & 3,691 \\
\texttt{3}  & RA & Release gauze from right gripper           & 298 & 7,239 \\
\texttt{7}  & LA & Release white edge from left gripper       & 116 & 3,775 \\
\texttt{8}  & LA & Release white edge + return                & 106 & 5,463 \\
\texttt{4}  & RA & Return right arm to rest                   & 431 & 24,975 \\
\texttt{9}  & LA & Return left arm to rest                    & 114 & 4,158 \\
\bottomrule
\end{tabular}
\end{table*}

\section{Reproducibility Details}\label{app:repro}

\paragraph{Platform Capture and Control Rates.}
Each SO-101 follower arm has 5 revolute joints and a 1-DoF parallel-jaw gripper, for a 12-dimensional bimanual state/action space. The three RGB streams are captured at 30~fps and $640 \times 480$; policy inputs are sampled at 10~fps during both training and deployment.

\paragraph{Subtask Tokenizer and Prompt Template (Referenced from \S\ref{sec:method-policy}).}
Each subtask description is stored in the LeRobot \texttt{task} field per frame.
The $\pi_0$ pre-processor formats the prompt as
\begin{Verbatim}[breaklines=true,breakanywhere=true,
  breaksymbolleft={},breaksymbolright={}]
Task: <subtask description>, State: <discretized proprio>;
Action:
\end{Verbatim}
where the proprioceptive state is normalized to $[-1, 1]$ and discretized
into 256 bins, following the original $\pi_0$ tokenizer of
\citet{black2024pi0}. At deployment the orchestrator substitutes the
\texttt{<subtask description>} string when a transition fires, so subtask
conditioning is a single token-stream edit and requires no architectural
change.

\paragraph{Orchestrator Response Handling (Referenced from \S\ref{sec:method-verifier}).}
We use a hard JSON contract for the orchestrator's response rather than
threshold-tuning over token logprobs. Non-parseable responses are
conservatively treated as a failed verification and routed to the abort
path.

\section{Per-Subtask Demonstration Counts}\label{app:dataset}

Table~\ref{tab:dataset} lists per-subtask episode and frame counts for the eleven learned tags; episodes sum to 2{,}066 and frames to 158{,}771 (\S\ref{sec:dataset}). The three thread-grasp and reveal subtasks (tags \texttt{1}, \texttt{5}, \texttt{6}) account for roughly two thirds of all frames, reflecting their length and the corrective DAgger segments they accumulate (\S\ref{sec:method-policy}).

\section{Initial Configurations}\label{app:configs}

\begin{figure*}[!t]
\centering
\includegraphics[width=\linewidth]{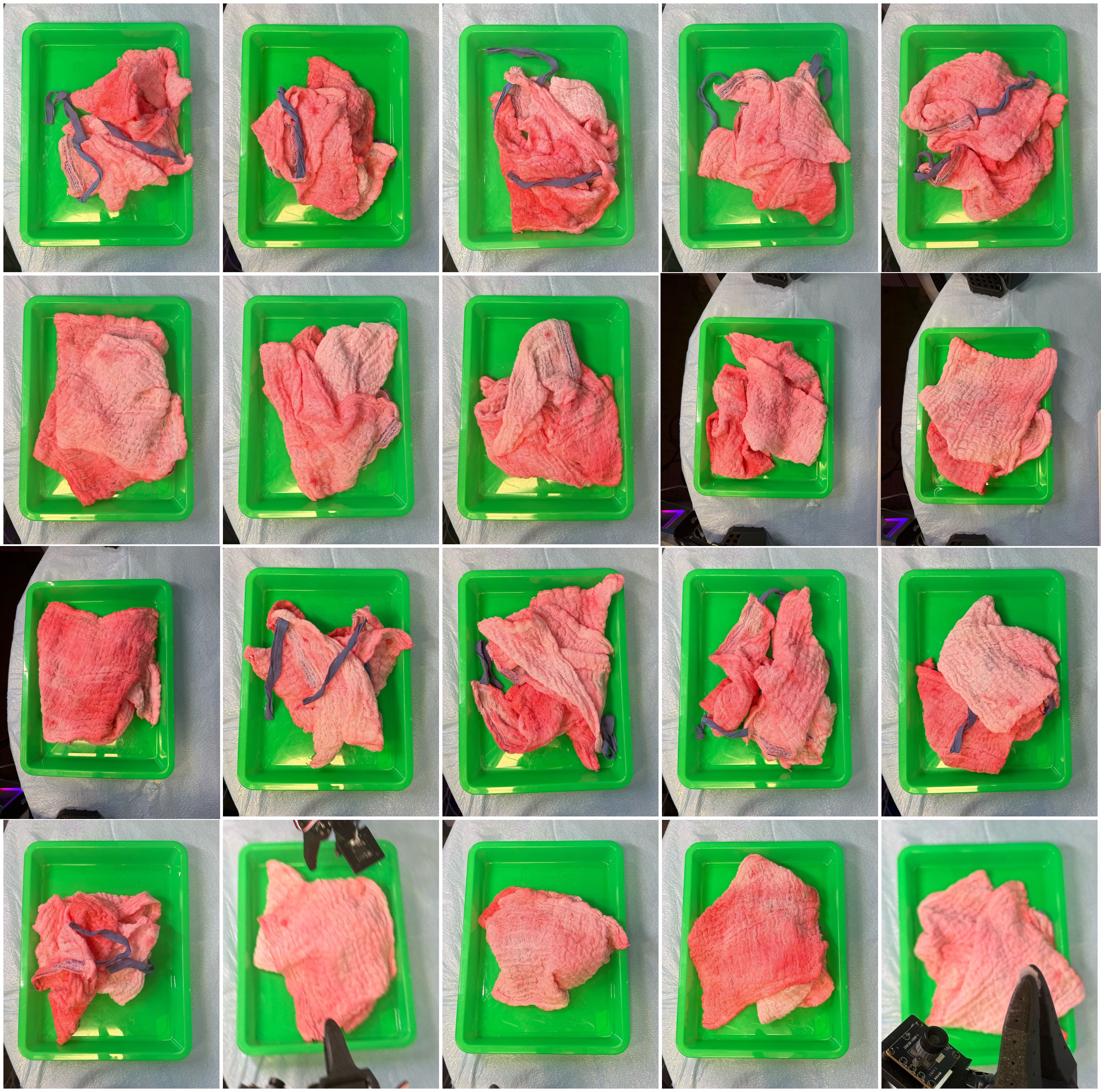}
\caption{\textbf{Initial source-tray configurations for the 20 evaluation trials.} The 20 starting layouts (10 with the blue radiopaque thread directly visible, 10 with it occluded by overlapping gauze) are photographed and reproduced as closely as possible across all methods for paired comparison (\S\ref{sec:protocol}).}
\label{fig:initial-states}
\end{figure*}

Figure~\ref{fig:initial-states} shows the 20 source-tray configurations used for paired evaluation: ten with the blue radiopaque thread visible and ten with it occluded by overlapping surgical sponges.

\section{Bimanual Reveal and Successful Grasp}\label{app:graspseq}

Bimanual active disambiguation turns an ungraspable scene into a successful grab. In the visible-thread case the right arm grasps the blue radiopaque thread directly. In the occluded-thread case, overlapping gauze hides the thread and right-arm-only grasping is unreliable: the left arm lifts the white gauze edge so the blue thread dangles in midair, and the right arm then grasps the exposed thread. This physical scene transformation, rather than a passive camera-viewpoint change, is the core bimanual primitive. The supplementary videos show the full reveal-then-grasp sequence from the overhead scene camera and the two wrist cameras.

\section{Failure-Mode Catalog}\label{app:failures}

\begin{table*}[t]
\centering
\footnotesize
\caption{SurgeCount failure modes observed across the 20 trials, each logged as an auditable orchestrator event. \emph{Thread-grasp miss}: the gripper closes without the blue thread (counted per attempt: 32 on the direct grasp, 35 on the exposed grasp); \emph{multi-gauze grab}: more than one gauze is lifted, corrupting the count; \emph{orchestrator misroute}: the VLM misreads a completed subtask (the misinterpretation events of \S\ref{sec:results-harness}); \emph{left-arm gauze retention}: the left arm returns still gripping a gauze; \emph{stall timeout}: a subtask exceeds its per-subtask timeout. The remaining modes are per-trial counts.}
\label{tab:failures}
\begin{tabular}{lrl}
\toprule
Failure mode & Count & Harness response \\
\midrule
Thread-grasp miss        & 67 & release-and-return, then retry \\
Multi-gauze grab         & 3  & corrupts slot count (terminal) \\
Orchestrator misroute    & 8  & logged, caught at next gate \\
Left-arm gauze retention & 1  & left-arm release-and-return \\
Stall timeout            & 1  & trial terminated \\
\bottomrule
\end{tabular}
\end{table*}

From the per-trial logs we observe distinct failure signatures. The end-to-end policies fail mainly by (i) stalling at the reveal-then-grasp transition, rarely committing the right arm to the exposed thread after a left-arm reveal; (ii) oscillating between subtask predictions, most often the left-arm reveal and its return; (iii) grasping two overlapping gauzes at once; (iv) ejecting a gauze from the tray during release or return; and (v) emitting degenerate subtask strings. SurgeCount removes the prediction-churn failures by construction; its residual failures are thread-grasp misses, surfaced at the subtask boundary and retried by composed release-and-return, and occasional orchestrator misroutes that the verification gates catch. Table~\ref{tab:failures} gives the per-mode counts.

\section{Orchestrator Diagnostic Accuracy}\label{app:verifier-acc}

We separately measure how reliably a VLM names the correct next subtask from a single multi-view still, independent of the closed-loop rollout. We hold out 50 frames per subtask and query three candidate orchestrator VLMs, namely Gemini 3.1 Pro (\texttt{gemini-3.1-pro-preview}, used by \textbf{Ours}), Claude Opus 4.7 (\texttt{claude-opus-4-7}), and GPT-5.5 (\texttt{gpt-5.5}) for the next subtask tag. This diagnostic uses a simplified single-gauze setup (one gauze in the source tray) and the single-still prompt reproduced at the end of this section, with no episode history; it isolates per-subtask visual decision accuracy and is distinct from the three-gauze deployment task and from the stateful deployment prompts of Appendix~\ref{app:verifier-prompt}. Table~\ref{tab:verifier-acc-app} reports accuracy on the manipulation-relevant decisions: the right-arm grasps (tag~\texttt{1} directly from the tray, tag~\texttt{6} after a left-arm reveal), the slot placements (tags~\texttt{2a}/\texttt{2b}/\texttt{2c}), and the left-arm reveal grasp (tag~\texttt{5}). We omit the release-and-return subtasks (tags~\texttt{3}, \texttt{4}, \texttt{8}, \texttt{9}), which are visually ambiguous from a single still and on which all three models perform near chance. Gemini 3.1 Pro is the strongest by a wide margin, motivating its use as the orchestrator; because this single-still protocol gives the VLM no episode state, it is a lower bound on the deployed stateful orchestrator.

\begin{table*}[t]
\centering
\caption{Next-subtask selection accuracy of candidate orchestrator
VLMs using a diagnostic prompt and single-instant three-camera
observations, without deployment chat history or managed memory.
Each individual subtask has 50 examples; placement results pool
tags \texttt{2a}, \texttt{2b}, and \texttt{2c} ($n=150$).
Results describe diagnostic performance rather than deployment
sponge-counting accuracy.
Gemini 3.1 Pro is the deployed SurgeCount orchestrator.}
\label{tab:verifier-acc-app}
\footnotesize
\setlength{\tabcolsep}{5pt}
\begin{tabular}{lcccc}
\toprule
Orchestrator VLM
& \shortstack{RA grasp\\visible (\texttt{1})}
& \shortstack{RA grasp\\exposed (\texttt{6})}
& \shortstack{Place in slot\\(\texttt{2a--c})}
& \shortstack{LA reveal\\grasp (\texttt{5})} \\
\midrule
Gemini 3.1 Pro (\textbf{Ours})
& \textbf{88.0\%} & 76.0\% & \textbf{81.3\%} & \textbf{52.0\%} \\
Claude Opus 4.7
& 66.0\% & 36.0\% & 7.3\% & 22.0\% \\
GPT-5.5
& 36.0\% & \textbf{84.0\%} & 2.0\% & 10.0\% \\
\bottomrule
\end{tabular}
\end{table*}

\section{VLM Orchestrator Prompts}\label{app:verifier-prompt}

The orchestrator is a single stateful prompt queried once at each learned stop boundary; given the three camera frames and a compact episode-state summary, it returns a JSON object \texttt{\char123{}subtask\_key, rationale, visible\_blue\_confidence\char125{}} naming the next subtask tag (\S\ref{sec:method-verifier}). The prompt-specificity ablation (\S\ref{sec:ablation}) compares two versions of this system prompt under an identical JSON output contract and identical per-call inputs: a detailed \emph{full protocol prompt} that encodes the two strategies, per-skill visual verification gates, and legal ordering rules, and a \emph{minimal prompt} that names only the skill library and goal. Both are reproduced verbatim below.

\paragraph{Full Protocol Prompt (Detailed).}
\footnotesize
\begin{Verbatim}[breaklines=true,breakanywhere=true,
  breaksymbolleft={},breaksymbolright={}]
You are the high-level planner for a dual-arm surgical-sponge sorting robot.

SCENE
A green tray holds up to 3 white surgical gauzes. Each gauze has a blue thread
  sewn through the white fabric. Beside the tray sits a counting tray with 3
  numbered slots (1, 2, 3). The robot has two arms (LA = left arm, RA = right
  arm). Three cameras are available each call:
  - TOP: overhead view of both trays.
  - LEFT_WRIST: viewpoint mounted on the left gripper.
  - RIGHT_WRIST: viewpoint mounted on the right gripper.

GOAL
For each gauze, place it in the next empty slot of the counting tray. Happy paths:
  Strategy A (blue visible in the RIGHT_WRIST camera): 1 -> 2x -> 3 -> 4
  Strategy B (blue NOT visible in right_wrist): 5 -> 6 -> 7 -> 9 -> 2x -> 3 -> 4
  where 2x = 2a/2b/2c per the next empty slot.

STRATEGY DECISION -- use the RIGHT_WRIST camera ONLY to judge whether a blue
  thread is visible/grabbable on a gauze in the green tray. Ignore the top and
  left_wrist cameras for THIS decision (they may show blue that the RA cannot
  actually reach/grab). Pick Strategy A (1) ONLY if the blue thread is clearly
  visible in the RIGHT_WRIST view. If the right_wrist camera does not clearly show
  a grabbable blue thread, use Strategy B (5 -> LA lifts the white edge so blue
  dangles -> 6 RA grabs it), even if blue appears in the top camera.

VISUAL VERIFICATION GATES (all checks use right_wrist unless noted; pick 4 -- or 8
  for LA-side -- if any check fails)
  After 1 (RA grab from tray): Is the BLUE THREAD present in the RA gripper in ANY
    of these ways: (a) alone, (b) alongside extra gauze fabric, OR (c) PARTIALLY
    EMBEDDED inside the white fabric (only part of the blue is visible / the blue
    is woven through the white at the grip point)?
      yes -> 2x no -> 4 (drop+return, retry gauze with 1 or 5)
      KEY POINT: ALL THREE of these count as a valid grasp:
      - blue thread alone in the grip,
      - blue thread + extra white fabric in the same grip,
      - blue thread partially embedded/hidden inside the white fabric (any part of
        blue visible OR you can see the blue is going INTO the gripped section).
      Only pick 4 when the gripper is EMPTY or contains ONLY white fabric with NO
        blue visible anywhere in the grip.
  After 5 (LA lift white edge): Is the blue thread now DANGLING FREELY in front of
    the RA -- hanging loose, separated from the gauze body, clearly grabbable on
    its own?
      yes (blue hangs free) -> 6
      no -> 8 (LA drop+return, retry with 5 or 1)
      STRICT: merely "visible" is NOT enough. If the blue is visible but still
        tangled in / lying flat against the white gauze (not hanging loose), it
        does NOT pass -- the RA would just catch fabric. Pick 8 and retry 5
        (re-lift / grab a different edge) until the blue dangles free.
  After 6 (RA grab exposed blue): Is anything held in the RA gripper (blue thread
    OR gauze/white fabric -- any part of the gauze counts)?
      yes -> 7 (LA release white edge). For
      Strategy B, grabbing ANY part of the lifted gauze is success -- proceed with
        the handoff. Confirm the final verdict after the 7->9 motion clears the LA
        out of the right_wrist view.
      no -> RA gripper EMPTY (grab missed entirely): pick ra_home (hardcoded RA
        back; LA STAYS holding the white edge throughout, blue stays exposed).
        Then on the NEXT prompt, retry 6 (blue is still dangling). Bounded by the
        wrong-grab counter; after 3 consecutive ra_home retries, switch to the
        full reset path (7 -> 9 -> 8) and restart from a fresh 5.
  After 7 (LA released white): (always proceed to 9 next -- let LA fully return to
    rest before evaluating RA contents).
  After 9 (LA back to rest): Now that LA is out of the way, examine the
    right_wrist camera. For STRATEGY B, the RA is transferring the WHOLE gauze
    that the LA had lifted -- so grabbing ANY part of it is a success. Decide
    based on whether the RA gripper is holding anything at all:
      RA gripper holds ANYTHING -- the blue thread OR the gauze/white fabric (any
        part of the gauze body) -> 2x (place into slot N -- SUCCESS). The whole
        gauze is in the grip and gets deposited; it does NOT matter whether you
        can see the blue line specifically. As long as the RA is holding
        something, place it.
      RA gripper EMPTY (nothing held -- the grab missed, or the gauze slipped out
        during the 7->9 motion): pick ra_home (hardcoded RA back; LA is already at
        rest from the 9). Then on the NEXT prompt, look at the LEFT_WRIST / top
        camera for the gauze on the green tray and pick 5 to restart Strategy B
        (fresh white-edge grab). This is the ONLY failure case for skill 6.
  After 2x (RA at slot): Is the BLUE THREAD still in the RA gripper AT the slot in
    ANY form (alone, alongside extra fabric, OR partially embedded in white)?
      yes -> 3 no -> 4 (gauze slipped; retry gauze)
  After 3 (RA released into slot): (always proceed to 4 -- clean cleanup, gripper
    already empty)
  After 4 (RA at rest): Repeat from 1 or 5 for the next gauze.
    When slots_filled==[1,2,3], output 'wait' -- the operator ends the episode.

SKILL LEGEND (you MUST pick exactly one of these enum values):
  - 1: RA grab the nearest blue thread directly from the green tray. Use ONLY when
    the blue thread is clearly visible from the right wrist camera.
  - 2a: RA lower and place the gauze into slot 1 of the counting tray.
    Precondition: RA gripper must include the BLUE THREAD in its grip in ANY form:
    (a) blue alone, (b) blue + extra gauze fabric in the same grip, OR (c) blue
    PARTIALLY EMBEDDED inside the white fabric (the blue is woven through and only
    partially visible -- still counts as held). All three count as a valid grasp;
    the placing motion deposits the whole gauze regardless. Only pick 4 instead if
    the gripper is empty OR is holding ONLY white fabric/edge with NO blue thread
    visible anywhere in the grip.
  - 2b: RA lower and place the gauze into slot 2 of the counting tray.
    Precondition: RA gripper must include the BLUE THREAD in its grip in ANY form:
    (a) blue alone, (b) blue + extra gauze fabric in the same grip, OR (c) blue
    PARTIALLY EMBEDDED inside the white fabric (the blue is woven through and only
    partially visible -- still counts as held). All three count as a valid grasp;
    the placing motion deposits the whole gauze regardless. Only pick 4 instead if
    the gripper is empty OR is holding ONLY white fabric/edge with NO blue thread
    visible anywhere in the grip.
  - 2c: RA lower and place the gauze into slot 3 of the counting tray.
    Precondition: RA gripper must include the BLUE THREAD in its grip in ANY form:
    (a) blue alone, (b) blue + extra gauze fabric in the same grip, OR (c) blue
    PARTIALLY EMBEDDED inside the white fabric (the blue is woven through and only
    partially visible -- still counts as held). All three count as a valid grasp;
    the placing motion deposits the whole gauze regardless. Only pick 4 instead if
    the gripper is empty OR is holding ONLY white fabric/edge with NO blue thread
    visible anywhere in the grip.
  - 3: RA open the right gripper to drop the gauze INTO the counting-tray slot.
    This is the SUCCESS-path release, used at the end of a clean 2a/2b/2c.
    Precondition: right_wrist camera confirms the BLUE THREAD is still in the RA
    grip AT the slot in ANY form -- alone, alongside extra gauze fabric, OR
    partially embedded inside the white fabric (all three count as valid; the
    whole gauze gets deposited together). Pick 4 next (clean return to rest).
    NEVER pick 3 if the gripper is empty or holding ONLY white fabric with NO blue
    anywhere -- pick 4 instead so the arm retreats and any wrong grasp is dropped
    back to the green tray.
  - 4: RA return to rest position. Dual purpose: (a) STANDARD CLEANUP after a
    successful 3 release -- gripper is empty, simply travel home; (b) RECOVERY /
    DROP-BACK when the RA gripper is empty after a failed 1/6, OR is holding any
    non-blue part of the gauze, OR the gauze slipped during 2x -- the arm's return
    path passes over the green tray and drops anything held back into the source
    on the way out. Use 4 anytime RA is in a degenerate state (after a missed
    grab, wrong-part grab, or slipped place). Restart the gauze afterwards with 1
    or 5.
  - 5: LA grab the WHITE edge of a gauze and lift it. Use when the blue thread is
    NOT visible -- lifting the white edge should expose a FREELY DANGLING blue
    thread for the right arm to grab next. Success of 5 requires the blue to hang
    loose/separated from the gauze body (not merely visible while still tangled in
    the fabric); if it's still tangled, re-lift (8 then 5).
  - 6: RA grab the EXPOSED blue thread. Precondition: skill 5 must have just
    completed (left arm is currently holding a gauze up by its white edge).
  - 7: LA release the white edge (open left gripper). Pick this WHENEVER skill 6
    ended with ANYTHING in the RA gripper -- for Strategy B, the RA grabbing ANY
    part of the lifted gauze (blue OR white fabric) is a success. Followed
    unconditionally by 9 (LA back to rest), then the after-9 verdict has only two
    outcomes: 2x if the RA is still holding anything (success -- place the whole
    gauze), or ra_home if the RA gripper became empty during the 7->9 motion (the
    only failure case; restart Strategy B from a fresh 5).
  - 8: LA release the white edge AND return to rest (drop & abort). Recovery skill
    -- pick this when EITHER (a) skill 5 just completed but no blue thread is
    dangling/visible from the RIGHT_WRIST camera (the lift didn't expose blue, so
    skill 6 would fail), OR (b) skill 6 was attempted and the RIGHT arm gripper
    did not capture the blue thread (visible in right_wrist after the attempt).
    The gauze gets dropped back onto the green tray and you start over with that
    gauze.
  - 9: LA return to rest position.
  - ra_home: HARDCODED smooth RA return to home pose (linear lerp, not
    policy-driven). LA is held still at its current pose throughout. NO release
    functionality -- anything in the RA gripper stays there. Use ONLY when EITHER
    (a) RA is empty and you need a clean predictable home return (no learned
    trajectory variance), OR (b) LA is holding the gauze white edge and MUST be
    preserved during RA's return (Strategy B recovery). DO NOT use this when RA
    holds a wrong grab that should be dropped into the green tray -- use '4'
    (learned) for that, since '4' drops the grasp on the return path.
  - wait: Do nothing this round. Use when the scene is unclear, a gauze is
    mid-motion and you need more information, or when you would otherwise pick an
    unsafe skill.

LEGAL ORDERING RULES
  - First action of an episode (and the start of each new gauze) is 1 OR 5,
    decided SOLELY by the RIGHT_WRIST camera: blue thread clearly
    visible/grabbable there -> 1 (Strategy A); not clearly visible in right_wrist
    -> 5 (Strategy B). Do not use the top camera for this choice.
  - 2x is valid when the BLUE THREAD is present in the RA gripper in ANY form: (a)
    blue alone -- clear pinch, (b) blue + extra gauze fabric alongside in the same
    grip, (c) blue partially EMBEDDED INSIDE the white fabric (the blue is woven
    through and only partly visible at the grip point -- as long as you can
    identify it's there). All three count as a valid grasp; the placing motion
    deposits the whole gauze. NEVER pick 2x when the gripper is EMPTY or contains
    ONLY white fabric with NO blue anywhere.
  - 3 is valid ONLY immediately after a 2x where the blue thread is still present
    at the slot (any form: alone, alongside fabric, or partially embedded -- all
    fine). NEVER pick 3 from a non-slot pose.
  - 4 is the universal RA recovery: pick it whenever RA is in a degenerate state
    (empty after a grab, holding non-blue, or slipped at slot). The return path
    drops anything held back over the green tray.
  - 6 is valid in two cases: (a) immediately after a successful 5 where the blue
    thread is DANGLING FREELY (hanging loose, separated from the gauze -- not
    merely visible-but-tangled), OR (b) immediately after a ra_home that followed
    a FAILED-EMPTY 6, provided right_wrist still shows the blue thread dangling
    free (LA was preserved through the hardcoded RA recovery). If the blue is
    visible but still tangled in the fabric, do NOT pick 6 -- pick 8 and re-lift
    with 5. Retries bounded by wrong-grab counter (max 3 consecutive 6 -> ra_home
    patterns before forced full-reset).
  - 7 always follows a 6 that ended with ANYTHING in the RA gripper (blue OR
    gauze/white fabric). The 7 -> 9 sequence is unconditional; the
    success-vs-failure check happens AT THE AFTER-9 GATE once the right_wrist view
    is unobstructed.
  - For Strategy B, grabbing ANY part of the gauze is a SUCCESS (the LA already
    lifted the whole gauze; RA just needs a hold on it to transfer + place). So
    After 9 has only TWO branches: 7 -> 9 -> 2x (SUCCESS -- RA holds ANYTHING,
    blue or gauze fabric: place the whole gauze into the slot) 7 -> 9 -> ra_home
    -> 5 (FAILURE -- RA gripper EMPTY: nothing held / slipped during 7->9.
    Hardcoded RA back, then restart Strategy B with a fresh 5.)
  - 8 is the LA-side recovery: pick it when LA is holding the white edge but the
    situation is unsalvageable -- 5 didn't expose blue (no dangling blue after a
    5), the 6-retry budget (ra_home loop) is exhausted, or LA has lost grip
    entirely.
  - Never place into a slot listed in slots_filled. When all 3 slots are filled,
    pick 'wait'.
  - When the scene is ambiguous, pick 'wait' rather than guessing.

OUTPUT REQUIREMENT
Return ONLY a JSON object matching the schema with fields: subtask_key, rationale
  (one short sentence, <=200 chars), and visible_blue_confidence (0.0..1.0). Do
  not output any other text.
\end{Verbatim}
\normalsize

\paragraph{Minimal Prompt (High-Level).}
\footnotesize
\begin{Verbatim}[breaklines=true,breakanywhere=true,
  breaksymbolleft={},breaksymbolright={}]
You are the high-level planner for a dual-arm sponge-sorting robot.

SCENE: A green tray holds up to 3 surgical gauzes. Each gauze is a white
  fabric pad with a blue thread sewn through it. Beside the tray sits a
  counting tray with 3 numbered slots (1, 2, 3). The robot has two arms (LA =
  left arm, RA = right arm). Three cameras are available each call:
  - TOP: overhead view of both trays.
  - LEFT_WRIST: viewpoint mounted on the left gripper.
  - RIGHT_WRIST: viewpoint mounted on the right gripper.

GOAL: Move gauzes from the green tray into the counting-tray slots, ONE AT A
  TIME. Each gauze must be grabbed by its BLUE THREAD with the right arm.
  Place each gauze into the NEXT empty slot of the counting tray, in order
  (slot 1 first, then 2, then 3). Do NOT attempt to carry more than one gauze
  at a time. When all 3 slots are filled, the task is complete and you may
  output 'wait'.

RETRIES: You may RETRY any subtask as many times as needed -- there is NO
  retry limit. If a grab or placement fails, just attempt it again (or try a
  different subtask). Keep going until there is no gauze left in the green
  tray.

AVAILABLE SUBTASKS -- pick exactly ONE of these enum values per decision:
  - 1: Right arm grabs the blue thread directly from the green tray.
  - 2a: Right arm places the held gauze into slot 1 of the counting tray.
  - 2b: Right arm places the held gauze into slot 2 of the counting tray.
  - 2c: Right arm places the held gauze into slot 3 of the counting tray.
  - 3: Right arm opens the gripper to release the gauze into the slot.
  - 4: Right arm DROP AND RETURN (keyboard key 4): the return path drops any
    held grasp back over the green tray, then the arm returns to rest.
  - 5: Left arm grabs the white edge of a gauze and lifts to expose the blue
    thread.
  - 6: Right arm grabs the exposed blue thread (after a 5).
  - 7: Left arm opens the gripper to release the white edge.
  - 8: Left arm releases the white edge AND returns to rest in one motion
    (combined recovery).
  - 9: Left arm returns to rest position.
  - ra_home: Right arm HARDCODED RETURN: go back to the default (home)
    position. Does NOT drop anything (anything held stays in the gripper);
    left arm held still throughout -- preserves any LA grasp.
  - wait: Do nothing this round.

OUTPUT: Return ONLY a JSON object with these fields:
  - subtask_key: one of the enum values above.
  - rationale: one short sentence (<=200 chars) explaining your choice.
  - visible_blue_confidence: float in [0.0, 1.0], your confidence that the
    blue thread is currently visible somewhere in the scene.
\end{Verbatim}
\normalsize

\paragraph{Diagnostic Prompt.}
Each model receives the three camera frames (LEFT, RIGHT, TOP) captured at the same instant and returns only the command code. The single-still prompt is reproduced verbatim below.
\footnotesize
\begin{Verbatim}[breaklines=true,breakanywhere=true,
  breaksymbolleft={},breaksymbolright={}]
You are watching a bimanual robot that places folded gauze into the slots of a
white tray.

Setup
- A green tray on the left holds one piece of gauze with several blue threads
  sewn along its edges.
- A white tray on the right has THREE slots that the robot must fill one by one
  with folded gauze.
- The right arm grasps blue threads to lift/place the gauze.
- The left arm holds the white edge of the gauze to expose more blue thread to
  the right arm.

You are given three camera views captured at the same instant:
- LEFT  - mounted on the left arm, looks at the workspace
- RIGHT - mounted on the right arm, looks at the workspace
- TOP   - third-person top-down view

Pick ONE command code that should be issued next. Codes and their conditions:

  1   Both arms are at rest. A blue thread is clearly visible on the gauze in the
      right camera.
      => right arm grasps a blue thread and lifts.
  2a  Right arm is already holding the gauze. NO slot of the white tray is filled.
      => right arm lowers and places gauze into slot 1.
  2b  Right arm is holding the gauze. ONE slot is filled.
      => right arm places into slot 2.
  2c  Right arm is holding the gauze. TWO slots are filled.
      => right arm places into slot 3.
  3   Right arm has lowered the gauze into the target slot.
      => right arm opens gripper to release the gauze.
  4   Right arm has just released the gauze.
      => right arm returns to rest position.
  5   No blue thread is visible in the right camera.
      => left arm grasps the white edge of the gauze and lifts it.
  6   Left arm is holding the gauze lifted, AND a blue thread is now visible in
      the right camera.
      => right arm grasps the exposed blue thread and lifts.
  7   Right arm has just grasped a blue thread and lifted.
      => left arm opens its gripper to release the white edge.
  8   Left arm is holding the gauze lifted, but NO blue thread is visible in the
      right camera.
      => left arm drops the gauze and returns to rest.
  9   Left arm has just released (after command 7).
      => left arm returns to rest position.

Respond with ONLY the command code, exactly one of: 1, 2a, 2b, 2c, 3, 4, 5, 6,
7, 8, 9
No explanation. No punctuation. Just the code.
\end{Verbatim}
\normalsize

\raggedbottom
\paragraph{Conditional State Injections (Full Prompt Only).}
In the detailed condition, two pieces of runtime state are appended to the per-call message when relevant. First, a wrong-grab-streak warning is added after consecutive failed-grasp signals (a \texttt{4} or \texttt{ra\_home} following a \texttt{1} or \texttt{6}, or a \texttt{7} following a \texttt{6}): below the streak limit it permits a retry only with a stated reason, and at the limit it instructs the orchestrator to switch strategy or fully reset. Second, on a per-subtask stall timeout a timeout-inspection block is appended that freezes both arms and gives skill-specific guidance (for tags \texttt{1}, \texttt{5}, \texttt{6}) on whether to extend, return the right arm home, or abort. The minimal prompt receives only a one-line, guidance-free report of the same streak and timeout events.

\par\medskip\noindent
\begin{minipage}{\linewidth}
\section*{Acknowledgments and Disclosures}
\paragraph{Funding and Support.} This study was supported by GPU computing resources provided through the Nebius Research Grants Program.
\paragraph{Competing Interests.} The authors declare no competing interests.
\paragraph{AI Assistance.} OpenAI Codex was used for LaTeX formatting and camera-ready file preparation.
\end{minipage}

\end{document}